\documentclass[12pt, letterpaper]{article}
\usepackage[colorlinks=true, urlcolor=blue, citecolor=green]{hyperref}
\usepackage[width=15cm, left=3cm, right=3cm, top=2cm, bottom=2cm]{geometry}
\usepackage{graphicx}
\usepackage{amsmath}
\usepackage{subcaption}
\graphicspath{{images/}}
\usepackage{titling}
\usepackage[table]{xcolor}
\usepackage{float}

\usepackage{fancyhdr}
\fancypagestyle{plain}{
  \fancyhf{}
  \fancyhead[L]{QZhou-Embedding Technical Report}
  \fancyhead[R]{\includegraphics[height=11pt]{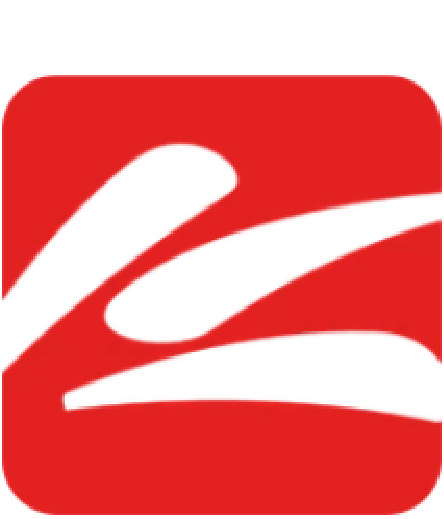} Kingsoft AI}
  
}

\pretitle{
  \vspace*{-1.8cm}
  \noindent\includegraphics[height=40pt]{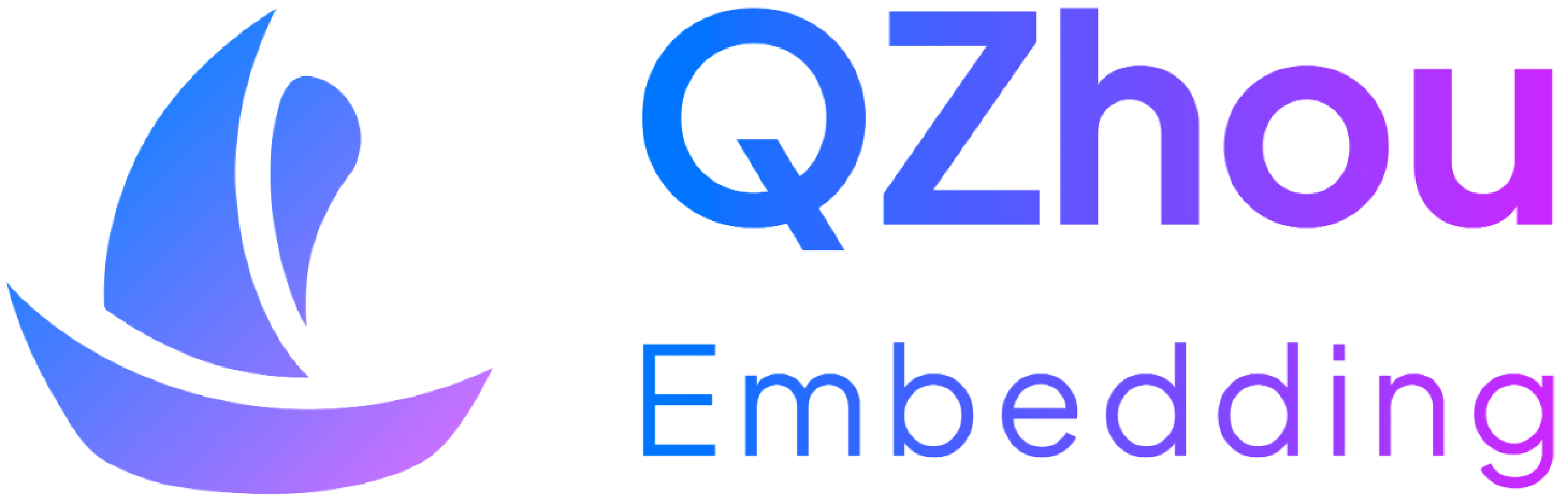}
  \begin{center}
    \noindent\rule{\textwidth}{4.0pt}\\[1em]
}
\posttitle{
  \noindent\rule{\textwidth}{0.2pt}
  \end{center}
}

\begin{document}
\title{\textbf{{\LARGE\bfseries QZhou-Embedding Technical Report}}}
\author{Peng Yu, En Xu, Bin Chen, Haibiao Chen, Yinfei Xu
\\\textbf{Kingsoft AI}\thanks{\href{https://kingsoft.com/}{https://kingsoft.com/}}}

\date{August 2025}

\maketitle

\begin{abstract}
We present QZhou-Embedding, a general-purpose contextual text embedding model with exceptional text representation capabilities. Built upon the Qwen2.5-7B-Instruct foundation model, we designed a unified multi-task framework comprising specialized data transformation and training strategies. The data transformation scheme enables the incorporation of more diverse textual training datasets, while the task-specific training strategies enhance model learning efficiency. We developed a data synthesis pipeline leveraging LLM API, incorporating techniques such as Paraphrasing, Augmentation, and Hard negative example generation to improve the semantic richness and sample difficulty of the training set. Additionally, we employ a two-stage training strategy, comprising initial retrieval-focused pretraining followed by full-task fine-tuning, enabling the embedding model to extend its capabilities based on robust retrieval performance. Our model achieves state-of-the-art results on the MTEB and CMTEB benchmarks, ranking first on both leaderboards(August 27, 2025), simultaneously achieves state-of-the-art performance on tasks including Reranking, Clustering, etc. Our findings demonstrate that higher-quality, more diverse data is crucial for advancing retrieval model performance, and that leveraging LLMs' generative capabilities can further optimize data quality for embedding model breakthroughs. Our model weights are released on HuggingFace\footnote{\href{https://huggingface.co/Kingsoft-LLM/QZhou-Embedding}{https://huggingface.co/Kingsoft-LLM/QZhou-Embedding}} under Apache 2.0 license. For reproducibility, we provide evaluation code and instructions on GitHub\footnote{\href{https://github.com/Kingsoft-LLM/QZhou-Embedding}{https://github.com/Kingsoft-LLM/QZhou-Embedding}}.

\end{abstract}

\section{Introduction}
Text embedding models, which transform natural language text into mathematical vector representations, play an indispensable role in text mining, question-answering systems, recommendation systems, and retrieval-augmented generation. Recently, LLM-based agent technology has experienced rapid development and widespread adoption, embedding models, which transform textual or multimodal data into vector representations for knowledge base construction, have significantly enhanced agent systems in terms of real-time performance, long-term memory, data privacy preservation, and knowledge integration capabilities. With the continuous advancement of neural networks and deep learning, text embeddings have evolved from early sparse representations (e.g., BM25\cite{ref1}) to dense representations based on fine-tuned deep networks such as BERT\cite{ref2} and T5\cite{ref3}, leading to significant performance improvements\cite{ref4}\cite{ref5}\cite{ref6}\cite{ref7}\cite{ref8}. In 2022, the rise of large language models (LLMs), exemplified by ChatGPT\cite{ref9}, ushered in a new era of text embeddings based on LLM representations, including models like text-embedding-3-large and RepLLaMA\cite{ref10}. Recent research on optimizing text embedding models has explored diverse perspectives and focal points. For instance, to address the limitation of decoder-only architectures—where causal attention mechanisms restrict token embeddings to unidirectional semantic capture—several approaches have been proposed: Echo Embedding\cite{ref11} employs input repetition and instruction design to enable preceding tokens to capture subsequent token semantics. LLM2Vec\cite{ref111} modifies attention to bi-directional mechanism to remove backward dependency constraints. Conan-Embedding-v2\cite{ref12} proposes a novel soft masking mechanism combined with dynamic rank reduction. Another widely adopted approach is knowledge distillation, where text embeddings are treated as the "signal states" representing textual semantics. By distilling knowledge from high-performing teacher models to student models, the objective is to optimize the embedding performance. For instance, Jasper\cite{ref13} employs a multi-stage knowledge distillation framework, combining with multiple carefully designed loss functions and finally achieving superior results. Debater\cite{ref15} proposes a step-by-step thinking mechanism for embedding generation, iteratively optimizing document representations through continuous COT. Distillation is applied to constrain the final token representation to learn the optimal semantic states from these thinking steps. Additionally, hard negative sampling has emerged as a crucial research direction in text embedding models, serving as a pivotal technique for model optimization. ANCE\cite{ref17} identified that conventional dense retrieval training leads to diminishing gradient norms during optimization. Thus they developed an asynchronous Approximate Nearest Neighbor (ANN) indexing mechanism that periodically refreshes the negative sample pool using the current model parameters, thereby ensuring the maintenance of up-to-date and optimally challenging negative samples. Both Conan-Embedding\cite{ref23} and its v2 version incorporated similar dynamic hard negative sampling techniques to enhance model performance. NV-Embed\cite{ref18} implemented an alternative approach by leveraging their previously developed NV-Retriever's\cite{ref19} positive-aware negative mining strategy, including TopK-MarginPos and TopKPercPos filtering mechanisms. \\\\
In this work, we present QZhou-Embedding, built upon the powerful Qwen2.5-7B-Instruct\cite{ref20} model, which pushes the boundaries of text embedding capabilities. To enhance the model's semantic understanding, we designed a unified multi-task learning framework that not only accommodates more diverse training data but also bring efficient learning across three key tasks: retrieval, natural language inference (NLI), and classification. Our framework comprises two core components: 1. Data Transformation: We carefully adapt data formats to the specific requirements of retrieval, NLI, and classification tasks, enabling effective feature extraction from heterogeneous data sources, significantly benefiting retrieval model training. 2. Training Strategy: We designed specialized loss functions based on each task's characteristics, optimizing model training efficiency. To further improve the robustness and generalization of vector representation, we propose a data synthesis method by employing three techniques to address data scarcity: Paraphrasing \& Data augmentation for limited datasets and Hard negative generation for negative sample enrichment. Building upon prior work, we designed a strategy named "Data Grouping Strategy", enabling batch sampling within single datasets, inadvertently increasing training difficulty through in-batch negative sampling from the same distribution. For model training, we used a two-phase training approach, through the first-stage retrieval training and second-stage full-capability training, our model acquires a solid foundation of retrieval capabilities, while effectively extending to multiple capability dimensions. Our model achieved state-of-the-art average scores on CMTEB\cite{ref21} and MTEB\cite{ref22} benchmarks, ranking first overall on both CMTEB and MTEB leaderboards, demonstrating the effectiveness of our approach. The contributions of our work are summarized as follows: 
\begin{itemize}
\item We propose a unified multi-task learning framework that systematically coordinates both data processing and training pipelines, enhancing diversity in datasets and efficiency in model training ;
\item We develop advanced data synthesis techniques powered by LLM, including Paraphrasing, Data augmentation, and Hard negative generation. These methods significantly enhance the quality of training corpora, thereby improving model's robustness and generalization capabilities;
\item We emply a two-stage training paradigm: Stage 1 focuses exclusively on retrieval capability building, establishing strong foundational retrieval performance; and stage 2 implements balanced training with controled retrieval/non-retrieval task ratios, achieving superior performance on classification (CLS), pair classification (PairCLS), and semantic textual similarity (STS) tasks while maintaining retrieval effectiveness;
\item Our model achieves state-of-the-art performance on both MTEB and CMTEB benchmarks, which validates the effectiveness of our proposed methods.
\end{itemize}

\section{Related Works}
\subsection{Text Embedding Models}
Text vector representation is a fundamental research area in natural language processing (NLP) and serves as the cornerstone for language understanding. Early approaches relied on sparse vector representations, such as TF-IDF\cite{ref24}, BM25\cite{ref26}, and LSA\cite{ref27}. With the advent of pretrained language models, dense contextualized representations based on architectures like BERT\cite{ref2} and T5\cite{ref3} became widely studied and applied\cite{ref4}\cite{ref5}\cite{ref6}. In the era of large language models (LLMs), major advancements have led to the development of LLM-based embedding models, such as text-embedding-3-small/large (OpenAI), E5-Mistral-7B\cite{ref28}, SFR-Embedding-Mistral\cite{ref29}, SFR-Embedding-2R\cite{ref30}, GRITLM\cite{ref31}, LLM2Vec\cite{ref111}, RepLLaMA\cite{ref10}, BGE-en-icl\cite{ref32}, NV-Embed\cite{ref18}, gte-Qwen2-7B-Instruct\cite{ref33}, Qwen3-Embedding\cite{ref34}, etc. These models benefit from optimized LLM architectures—such as RoPE positional encoding\cite{ref35}, RMSNorm\cite{ref36}, and GeGLU activation\cite{ref37}—combined with their strong semantic contextualization capabilities acquired through large-scale pretraining. As a result, LLM-based embeddings achieve superior performance in retrieval and related tasks.

\subsection{Embedding Model Training}
The mainstream approaches currently involve contrastive learning pretraining on unsupervised/weakly supervised corpora and supervised contrastive learning training on high-quality labeled positive and negative samples. In unsupervised learning, early work like SimCSE\cite{ref7} proposed feeding continuous inputs of both original and noise-augmented texts while employing contrastive learning to enhance the model's discriminative representation capability. For weakly supervised learning, gte\cite{ref33} utilized large-scale structured data (web search data, title-article pairs, etc.) for pretraining, followed by fine-tuning on high-quality open-source retrieval training data, achieving performance comparable to OpenAI embeddings with significantly fewer parameters. Conan-Embedding\cite{ref23} and v2 similarly adopted the weakly supervised pretraining \& supervised fine-tuning approach but incorporated techniques like cross-GPU batch loss balancing, dynamic hard negative mining, and soft masking (v2) to optimize the model. Seed1.6-Embedding\cite{ref38} employed a phased training strategy combining text and multimodal pretraining followed by business-scenario-specific fine-tuning, achieving superior representation quality.\\\\
Substantial research has also been conducted on modeling different tasks. Piccolo2\cite{ref39} introduced multi-task hybrid loss functions for diverse downstream tasks, an approach we also incorporate. SFR-Embedding\cite{ref30} utilized multi-task learning techniques to regularize embeddings, significantly enhancing domain data discrimination. Xiaobu-embedding unified the treatment of major CMTEB problem categories from the perspective of circle loss\cite{ref40}, fully leveraging multiple positive examples in original datasets while carefully balancing different loss weights.

\subsection{Data Synthesis}
Data quantity and quality are the most critical factors in model optimization, data synthesis methods have become a critical research direction due to the high cost of manual annotation. Doc2Query\cite{ref41} and Query2Doc\cite{ref42} employ question-answering models to generate pseudo-queries and pseudo-documents respectively, enhancing data for improved RAG performance. Promptagator\cite{ref43} addresses few-shot retrieval scenarios by generating queries of diverse intents using few-shot demonstrations and annotations, effectively improving retrieval capabilities across varying intents or distributions. GPL\cite{ref44} utilizes existing T5 encoder-decoder models to generate queries, retrieves similar passages as hard negatives using existing retrieval models, and employs cross-encoders to score each (query, passage) pair. Unnatural Instructions\cite{ref45} leverages prompt and in-context learning (ICL) techniques to generate synthetic examples through controlled instructions, inputs, and constraints, producing 64k diverse data entries from several seed examples with promising experimental results. Qwen3-Embedding\cite{ref34} designs a diversified prompting strategy by assigning document-specific roles to simulate potential
users querying that document, enabling LLMs to generate stylistically authentic queries that enhance diversity and realism.

\subsection{Hard Negative Mining Techniques}
Hard negatives serve as essential components in contrastive learning for retrieval model training. Early work like ANCE\cite{ref46} proposed an asynchronous ANN indexing mechanism that periodically updates hard negatives using checkpoint states to maintain optimally challenging samples. Conan-Embedding\cite{ref23} and its v2 version implemented a dynamic hard negative sampling strategy by excluding and refreshing samples when their scores fall below a threshold. NV-Retriever\cite{ref47}  proposed positive-aware negative mining, introducing TopK-MarginPos and TopKPercPos filtering criteria to minimize false negatives. LGAI-Embedding\cite{ref16} built upon NV-Retriever's strategy with adaptive margin-based mining strategies, employing ANNA IR as a teacher retrieval model to identify high-quality hard negatives while using TopKPercPos filtering to eliminate false negatives. 

\section{Unified Multi-task Learning Framework}
Embedding models support numerous downstream tasks including retrieval, reranking, STS, and classification. Given the diversity of these tasks and their associated data complexity, we explore a unified strategy to effectively handle them collectively while promoting optimization of the embedding model. Existing research on unified task processing includes circle loss\cite{ref40}, which approaches sentence pair similarity from a global perspective by categorizing tasks into class-level labels and pair-wise labels, Xiaobu-embedding demonstrated significant improvements by adopting this approach. Other models like Piccolo2\cite{ref39}, SFR-Embedding\cite{ref30}, NV-Embed\cite{ref47}, Conan-Embedding\cite{ref23} , and Conan-Embedding-v2 have incorporated multi-task learning using diverse training data with varying label processing methods, some employing task-specific losses (InfoNCE\cite{ref48}, Cosent\cite{ref49}, etc.). \\\\
Our design principle aims to accommodate more tasks and data types, enabling cross-domain and cross-task data to effectively enhance embedding capabilities. We propose a unified multi-task learning framework that categorizes training data into three task types: retrieval, NLI, and classification, with customized data and training solutions for each, allowing most natural text data to be converted into embedding training data through this framework. The following sections detail the framework's components and implementation methods.

\begin{figure}[htbp]
  \centering
  \includegraphics[width=0.6\textwidth]{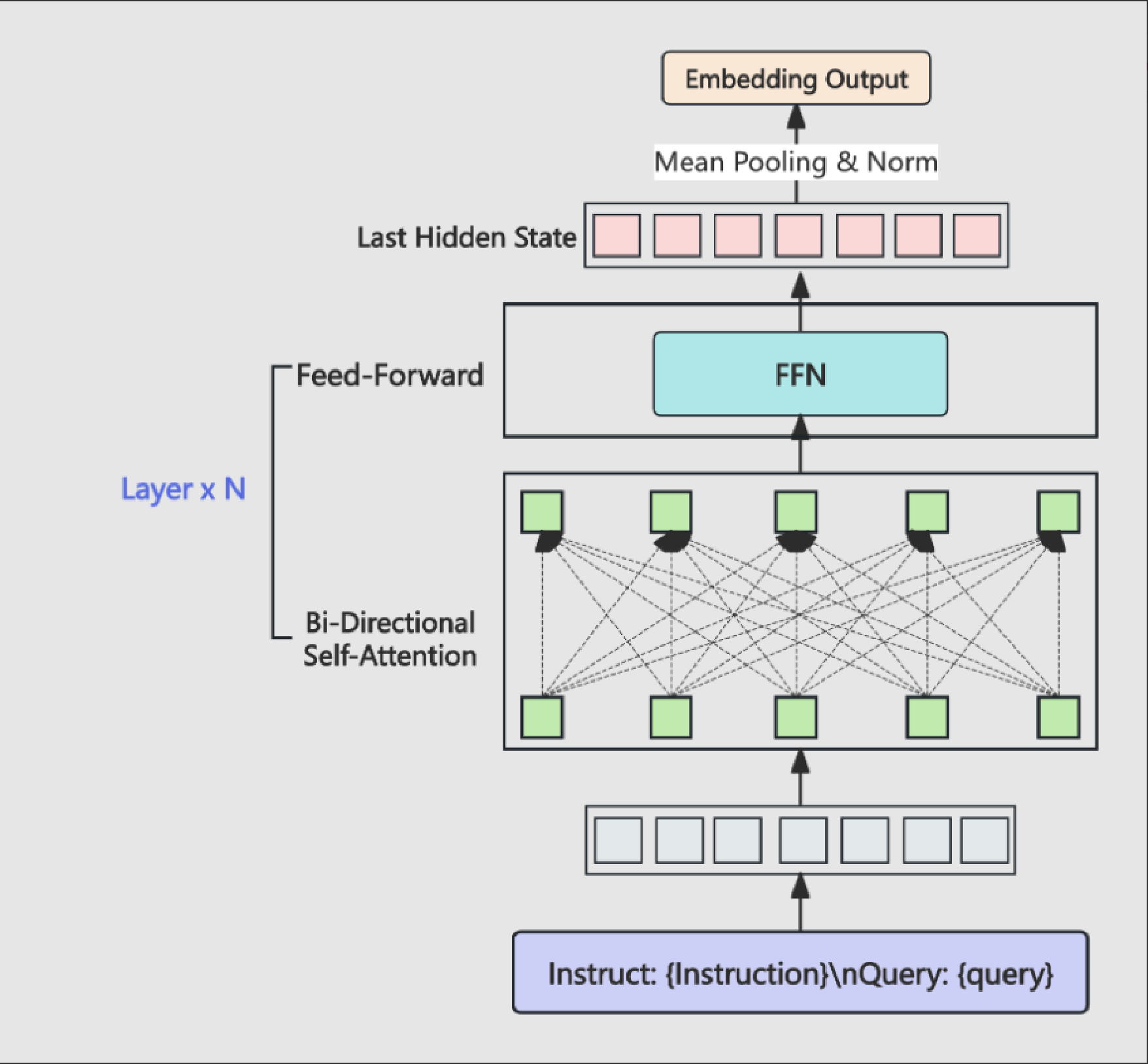} 
  \caption{QZhou-Embedding Architecture}
  \label{fig:f1}
\end{figure}

\subsection{Model Architecture}
Embedding models based on BERT or T5 \cite{ref39}\cite{ref14}\cite{ref50}\cite{ref23} exhibit powerful contextual representation capabilities, primarily attributed to their bidirectional attention mechanisms. However, recent large language models predominantly adopt decoder-only architectures with unidirectional attention, significantly constraining tokens' ability to capture contextual information. Several studies have addressed this limitation through architectural modifications or attention mechanism optimizations\cite{ref111}\cite{ref31}\cite{ref47}. Our work builds upon the Qwen2.5-7B-Instruct architecture and checkpoint due to its exceptional Chinese language contextual capabilities. Consequently, we implemented the following modifications: (1) modifying the original causal attention to bi-directional attention to enable comprehensive context capture, and (2) employing mean pooling with subsequent normalization to produce final embedding vectors. The model architecture is shown in Figure \ref{fig:f1}

\subsection{Data Transformation}
\label{chpt:c2}
\subsubsection{Retrieval-oriented Process}
While open-source datasets such as MS MARCO\cite{ref65} are readily accessible, they alone are insufficient for further advancing embedding model capabilities, thus we supplement with data from additional sources, such as news, academic paper and QA datasets. Given the heterogeneous nature of these datasets across domains and purposes, we design a retrieval-oriented data transformation methodology to convert diverse sources and formats into training data suitable for retrieval task. Below we outline selected categories of training data used for transformation and their processing procedures:
\begin{itemize}
\item \textbf{Title-Body/Abstract} "Title-Body/Abstract" type data primarily consists of title-body/article pairs typically sourced from online news, articles, documents, arXiv publications and Wikipedia. For these data types, the transformation process involves using the title as the query and the body/abstract as the positive sample. However, since the latter are documents, truncation is applied when they exceed the maximum training length.
\item \textbf{Claim-Evidence} This data type typically presents a claim or statement followed by extracted evidence that either supports or refutes it, commonly used for multi-hop fact extraction and claim verification tasks. Datasets generally contain claims and corresponding evidence, with each evidence instance labeled as "Supports" or "Refutes". The transformation process involves: converting the claim portion into a query sample, for evidence labeled as "Supports", the text is treated as a positive sample; for evidence labeled as "Refutes", it is converted into a negative sample.
\item \textbf{Question-Answer} Question-answering data and conversational Q-A pairs primarily originate from chat platforms and forums. Within the current wave of LLM and reinforcement learning research, such data exhibits remarkable volume and diversity. Virtually single-turn Q-A datasets(one question paired with one answer) represents the most suitable format for retrieval training. For transformation, the "Question/Query/User" portion is converted into queries, while the "Answer/Response/Assistant" portion is processed as documents.
\end{itemize} 

\subsubsection{NLI-oriented Process}
Natural Language Inference (NLI) represents a fundamental capability of NLP models, encompassing tasks such as semantic similarity, textual entailment, and sentiment analysis. This section describes the methodology for transforming and constructing training sets from NLI-style data, using textual semantic similarity (STS) and textual entailment tasks as illustrative examples. Our approach distinctively reformulates NLI tasks into text\_pair-score formats compatible with Cosent loss\cite{ref49} training strategy, where sample pairs are quantitatively scored based on their semantic relationships. The processing procedures for each are detailed below: 
\begin{itemize}
\item \textbf{STS} Semantic Textual Similarity (STS) is characterized by its symmetric semantic matching to determine whether two sentences share equivalent meaning. STS datasets typically consist of sentence pairs with associated labels, which may be binary classifications (yes/no, true/false) or numerical scores (e.g., 1.2, 3.1, 4.8). For binary labels, "yes"/"true" are mapped to a numerical value of 1, while "no"/"false" are converted to 0. The data is then structured into (query, document, score) triplets. Due to the symmetric nature of STS, each single original data sample can generate two training triplets by interchanging the query and positive document roles.
\item \textbf{Textual Entailment} Textual entailment further examines a model's capabilities in reasoning, typically featuring three-class labels: entailment, neutral, contradiction. Our processing method employs a three-tier scoring system: labels are assigned values of 2, 1, and 0 for entailment, neutral, and contradiction respectively. We construct (query, document, score) triplets accordingly, and similarly leverage symmetry to double the dataset size.
\end{itemize} 

\begin{figure}[htbp]
  \centering
  \includegraphics[width=\linewidth]{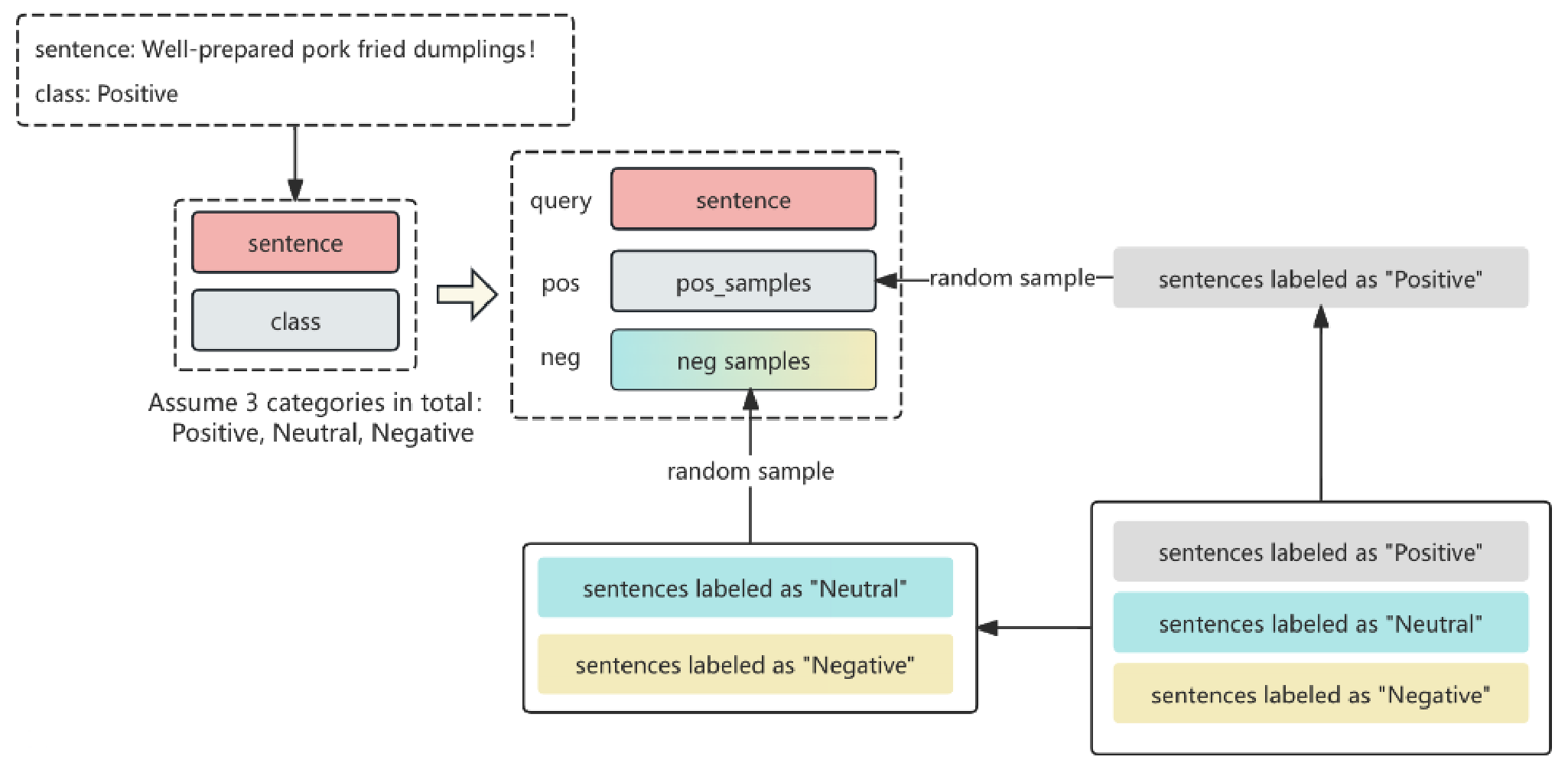}
  \caption{CLS-oriented data transformation}
  \label{fig:f4}
\end{figure}

\subsubsection{CLS-oriented Process}
Classification tasks encompass text categorization and sentiment classification scenarios, it typically follows a (text, label) format, where texts within the same category exhibit semantic proximity while distinct boundaries separate different classes. NV-Embed\cite{ref47} compared label-based and example-based data construction methods, with experimental results demonstrating the superiority of the latter. Adopting the example-based approach, we process classification data (text, label) by using the text as query, sampling other texts sharing the same label as positive examples, and selecting texts from different labels as negative examples. Figure \ref{fig:f4} provides a detailed schematic illustration of this process.

\subsection{Training Strategy}
Each task category—retrieval, NLI, and classification—operates within a data construction process respectively, for which we have designed specialized training objectives to to enhance model training efficiency. This section elaborates on the design of loss functions for retrieval, NLI, and classification tasks.
\subsubsection{Retrieval}
For the retrieval task, we adopt the widely used InfoNCE loss\cite{ref48}, but incorporate an improvement inspired by gte\cite{ref33} by augmenting the original query-negative loss with an additional query-query loss term. Specifically, each query within a batch is treated as a negative sample for all other queries. The final loss formulation is explicitly described in Equation \eqref{eq:l1}. 

\begin{equation}
\mathcal{L}_{\text{Retrieval}} = -\frac{1}{n}\sum_i\log \frac{e^{\text{sim}(q_i, d_i^+) / \tau}}{e^{\text{sim}(q_i, d_i^+) / \tau}+
\sum_{j}e^{\text{sim}(q_i, d_j^-) / \tau} + 
\sum_{j \neq i} e^{\text{sim}(q_i, q_j) / \tau}
}
\label{eq:l1}
\end{equation}

\subsubsection{NLI}
For NLI tasks, the transformed labels are numerically comparable and exhibit ordinal relationships. We employ Cosent loss\cite{ref49} to optimize such data, which is designed based on the principles of Circle loss\cite{ref40}. As a ranking-sensitive loss function, Cosent loss requires only ordinal label information for optimization while demonstrating faster convergence. Its mathematical formulation is presented in Equation \eqref{eq:l2}.

\begin{equation}
\mathcal{L}_{\text{NLI}} = \log (1+\sum_{\text{sim}(i, j)>\text{sim}(k, l)}exp(\frac{\text{sim}(x_k, x_l)-\text{sim}(x_i, x_j)}{\tau}))
\label{eq:l2}
\end{equation}

\subsubsection{CLS}
The classification loss also adopts the InfoNCE objective. However, since CLS data is processed in an example-based manner, directly applying in-batch negative sampling on classification datasets with limited categories may lead to false negatives from items of different classes. Numerous studies have proposed diverse approaches to address this issue\cite{ref51}\cite{ref52}\cite{ref47}. We propose a masking mechanism that appends class labels to each positive and negative sample during preprocessing (recorded as separate variables rather than modifying raw text). During in-batch negative sampling, for each negative sample from other data instances, we check whether its label matches the current query’s class. If matched, the negative loss contribution is masked to zero to prevent erroneous penalization; otherwise, it is normally computed. The core loss remains InfoNCE, with the CLS loss formulation shown in Equation \eqref{eq:l3}. Where $C_{t_i}$ denotes the class label of sample $t_i$, and $n$represents the number of negative samples per data instance.

\begin{equation}
L_{\text{CLS}} = -\frac{1}{n}\sum_i \log \frac{
    e^{\text{sim}(t_i, t_i^+)/\tau}
}{Z_i}
\label{eq:l3}
\end{equation}

\begin{align*}
\text{where} \ Z_i=& \left. e^{\text{sim}(t_i, t_i^+)/\tau} + \sum_n \text{MASK}(t_i, t_{i,n}^-) \cdot e^{\text{sim}(t_i, t_{i,n}^-)/\tau} + \right. \\
& \left. \sum_{j \neq i} \text{MASK}(t_i, t_j) \cdot e^{\text{sim}(t_i, t_j)/\tau} + \right. \\
& \left. \sum_{j \neq i}\sum_n\text{MASK}(t_i, t_{j,n}^-) \cdot e^{\text{sim}(t_i, t_{j,n}^-)/\tau} \right.
\end{align*}

\begin{align*}
\text{and} \ C_{t_i} = C_{t_i^+}
\end{align*}

\begin{align*}
\text{and} \ \text{MASK}(t_i, t_j) = 
\begin{cases}
0 & \text{if } C_{t_i} = C_{t_j}, \\
1 & \text{otherwise}
\end{cases}
\end{align*}

\section{Data Synthesis}
The production of higher-quality data through data production has gained critical importance in embedding training. Manual annotation incurs higher costs and lower production efficiency, thus developing effective automated data synthesis methods has emerged as a key research focus. Recent advancements in large language models (LLMs) have significantly improved their linguistic capabilities, enabling accurate interpretation of human instructions and generation of high-quality outputs. Multiple existing methods have effectively leveraged LLMs to generate high-quality data\cite{ref28}\cite{ref34}, we similarly leverages LLM capabilities for data production across three dimensions: structural diversity, semantic diversity, and difficulty, with dedicated synthesis strategies for each. For structural diversity, we propose Paraphrasing techniques; for semantic diversity, we introduce Augmentation methods; and to increase training difficulty and improve semantic discriminability, we employ LLMs to generate more challenging hard negative examples. The following sections detail these methodologies. The constraint components for all data synthesis techniques are specified in Table \ref{tab:t1} of Appendix \ref{app:a1}.
\subsection{Structural Diversity Enhancement}
\begin{figure}[htbp]
  \centering
  \includegraphics[width=\linewidth]{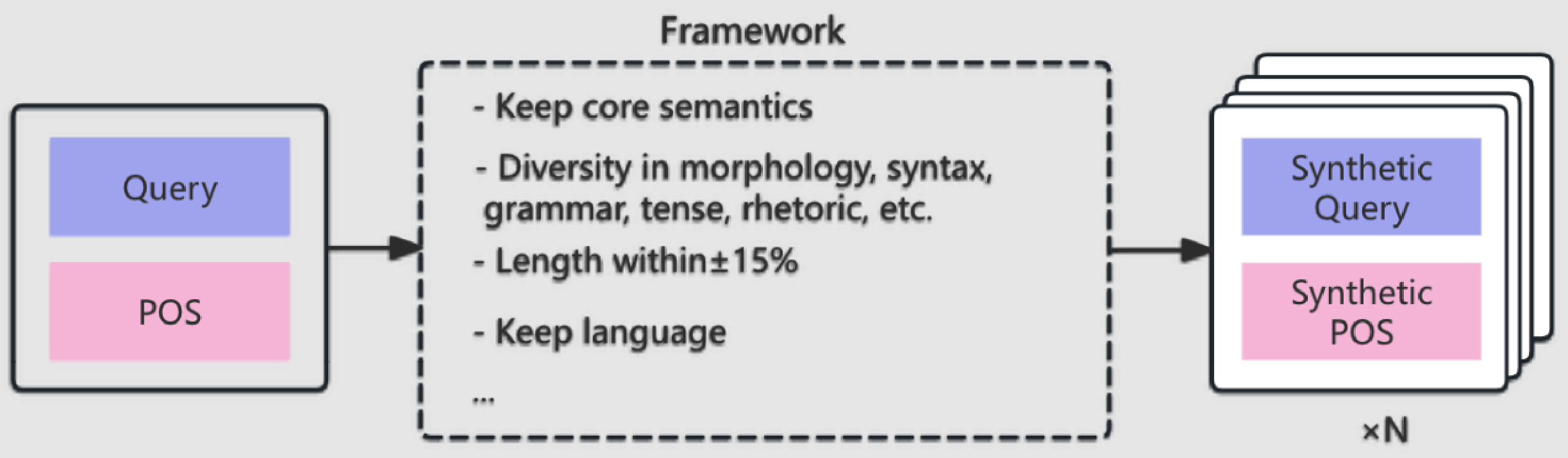}
  \caption{LLM-based Paraphrasing Workflow}
  \label{fig:f5}
\end{figure}

Linguistic structures of text encompass lexical, syntactic, and grammatical features, which represent relatively surface-level characteristics reflecting word arrangements, combinations, tenses, voices, and other formal attributes. Embedding models must accurately capture underlying semantics despite variations in surface form, ensuring robustness to external structural changes. For example, the following two sentences, despite structural differences, should be recognized as semantically equivalent:
\begin{itemize}
\item The cat chased the mouse.
\item The mouse was chased by the cat.
\end{itemize} 
To effectively train an embedding model that remains invariant to structural variations while accurately capturing semantic information, we propose a Paraphrasing strategy. For each training sample containing a query and a positive document, we apply LLM-based paraphrasing to both contents, generating augmented instances that preserve semantic equivalence while introducing structural divergence. The prompt constraints and workflow are illustrated in Figure \ref{fig:f5}. 

\subsection{Semantic Diversity Enhancement}
\begin{figure}[htbp]
  \centering
  \includegraphics[width=\linewidth]{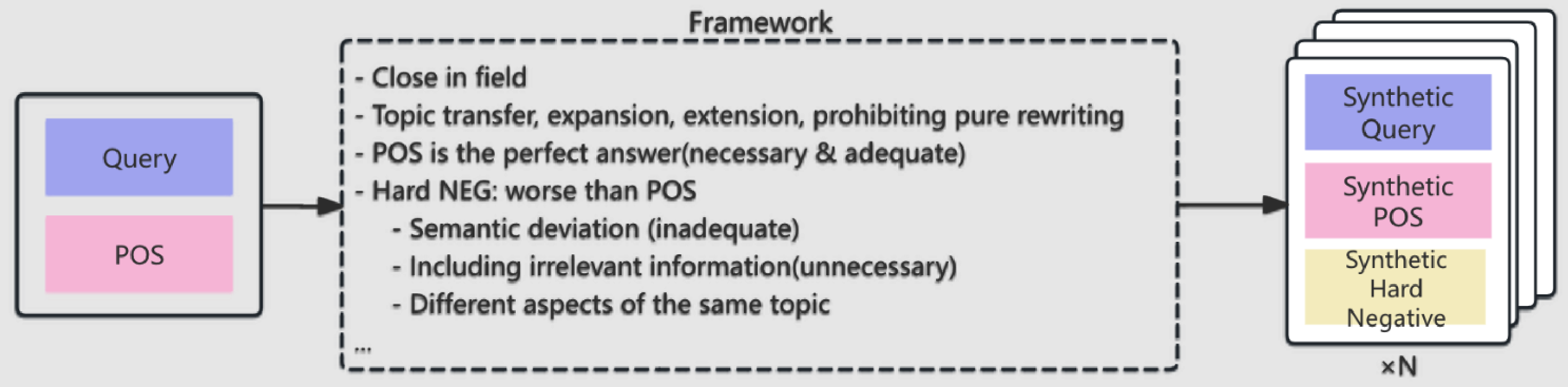}
  \caption{Semantic Augmentation Workflow}
  \label{fig:f6}
\end{figure}

Merely augmenting data through superficial structural modifications yields negligible improvements in model capabilities, as generalization relies not only on structural disentanglement but also on diverse topics and content to ensure uniform vector representations in the spatial domain. Therefore, beyond paraphrasing, we propose an augmentation method using LLM to diversify semantics. The core concept is: given a complete (query, positive) pair, the model must comprehend the domain and perspective discussed and learn to expand into different topics, aspects, and viewpoints while remaining contextually anchored. This process is governed via prompt constraints. The Augmentation framework is illustrated in Figure \ref{fig:f6}.

\begin{figure}[htbp]
  \centering
  \includegraphics[width=\linewidth]{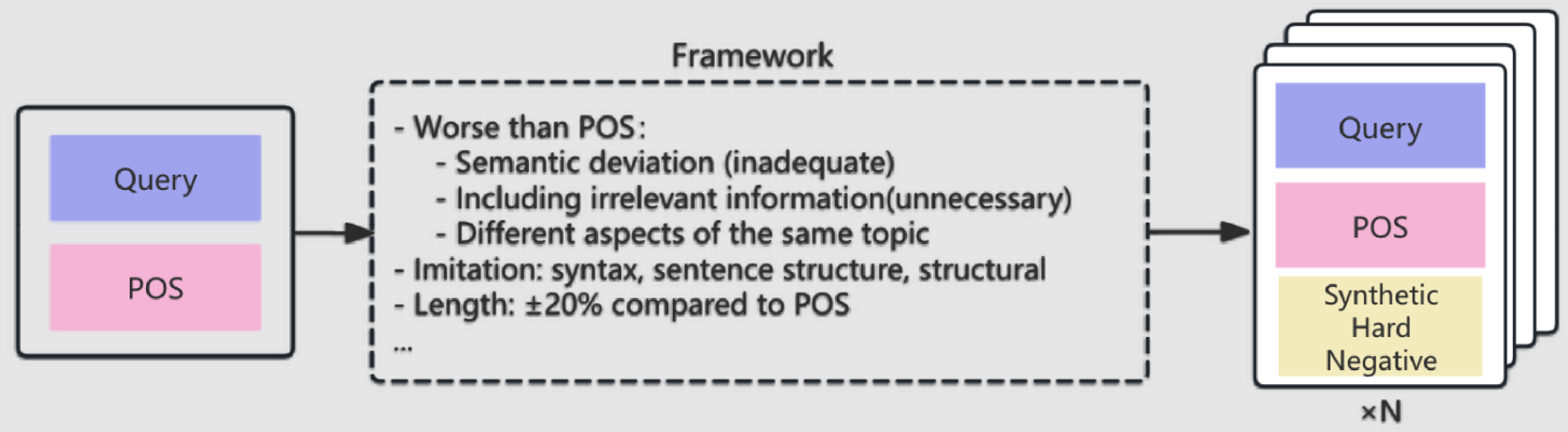}
  \caption{Hard Negative Synthesis  Workflow}
  \label{fig:f7}
\end{figure}

\subsection{More challenging embeddings}
Hard negative examples are crucial for enhancing the performance of text embedding models, often requiring substantial effort to acquire. Leveraging the linguistic capabilities of large language models, we design an automated hard negative synthesis method tailored for retrieval datasets. Our domain-specific experiments demonstrate that large language models can generate examples that are indistinguishable, the framework is illustrated in Figure \ref{fig:f7}. \\\\
During Data paraphrasing and Augmentation, we implement task-specific strategies: for retrieval tasks, we rewrite/expand (query, positive) pairs and add them to the original dataset; for NLI tasks, we rewrite individual sentences by randomly duplicating existing entries containing the original sentences and replacing them with rewritten versions to achieve data expansion—without applying augmentation to prevent ambiguity; for classification tasks, we rewrite sentences while retaining their original labels, example-based processing was applied using the rewritten results, again without employing augmentation. We provide several data synthesis examples in Appendix \ref{app:a3} for reference.

\begin{figure}[htbp]
  \centering
  \includegraphics[width=\linewidth]{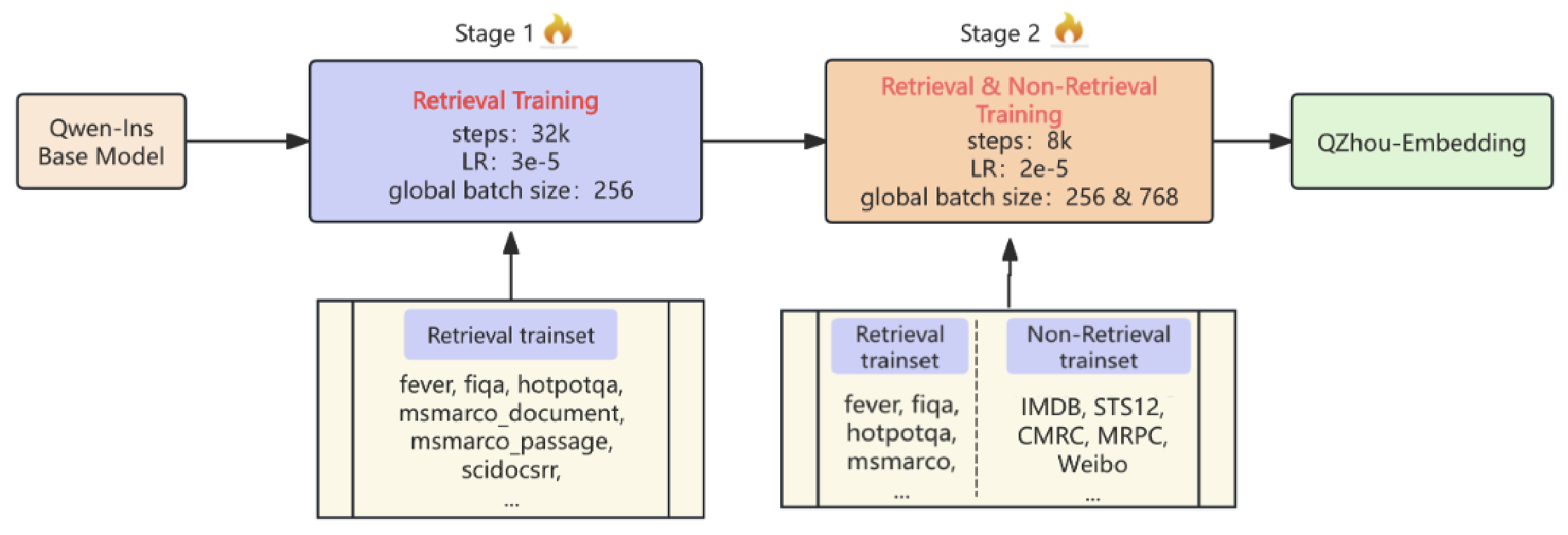}
  \caption{Training pipeline}
  \label{fig:f8}
\end{figure}

\section{Training Optimization}
\subsection{Data Grouping Strategy}
Prior works like Linq-Embedding\cite{ref52} and SFR-Embedding-Mistral\cite{ref30} adopted task-homogeneous batching, partitioning data by task rather than mixing them, and sampling tasks based on weighted randomness during training. Building on this, we propose a refined Data Grouping Strategy, extending the granularity from task-level to dataset-level partitioning. We posit that dataset-level grouping captures more domain-specific clustering patterns—samples within the same dataset often exhibit inherent domain similarities, while such consistency may not hold across datasets. \\\\
Our approach partitions training data into subsets by name. During training, only samples from a single dataset are sampled per batch, with file pointers recorded to enable sequential reading in subsequent iterations. For sampling weights, we adopt the data sampling strategy from gte\cite{ref33} and mgte\cite{ref50}, scaling weights by dataset size followed by normalization. For dataset $i$ with size $l_i$, its sampling weight is computed as Equation \eqref{eq:l4}

\begin{equation}
p_i=\frac{l^\alpha_i}{\sum_{j=1}^{m}l_j^\alpha}
\label{eq:l4}
\end{equation}

\subsection{Two-Stage Training}
Inspired by NV-Embed's\cite{ref47} two-stage contrastive learning instruction tuning technique, we adopt a similar training approach: the first stage exclusively uses retrieval-oriented training data, while the second stage integrates both retrieval and non-retrieval tasks, the overall training framework is illustrated in the figure \ref{fig:f8}. Two key distinctions are incorporated: first, we integrate the previously described Data Grouping Strategy; second, we implement global control over the sampling ratio of retrieval training datasets, since our findings indicate that naively incorporating additional data significantly degrades retrieval performance. \\\\
For global control of sampling ratio, a hyperparameter $\eta$ is introduced into the sampling function to control the proportion of retrieval training, ensuring that throughout the second training stage, the computational contribution of retrieval data accounts for $\eta$, while non-retrieval data constitutes $1-\eta$. The following set of equations formalizes the computational process from partitioned datasets to sampling ratio determination. Let the training data $D=[d_1, d_2, ..., d_N]$ , where each $d_i$ represents a distinct dataset (e.g., MSMARCO\_passage, SQUAD), with corresponding sizes $L=[l_1, l_2, ..., l_N]$. Following the aforementioned strategy, we first apply an exponential scaling factor $\alpha$, a mask factor $M$ is then applied to filter retrieval and non-retrieval training sets for summation. The equations are as follows:
\begin{align*}
S_{ret}& \left. =\sum_iM_i \cdot l^\alpha_i  \right. \\
S_{non\_ret}& \left. =\sum_i(1-M_i) \cdot l^\alpha_i \right. \\
where \ M_i & \left. = 
\begin{cases}
0 & \text{if } d_i \in \text{RET}, \\
1 & \text{else}
\end{cases} \right. 
\end{align*}
where \ $\text{RET}$ denotes the set of retrieval training datasets. The retrieval ratio is then scaled using $\eta$ to derive the final normalized sampling ratios for the training sets:
\begin{align*}
L_{samp}& \left. =[l_1^{samp}, l_2^{samp}, ...l_N^{samp}] \right.
\end{align*}
\begin{align*}
where \ l_i^{samp}& \left. =
\begin{cases}
\frac{\eta_{RET} \cdot l_i^\alpha}{S_{ret}} & \text{if } d_i \in \text{RET}, \\
\frac{(1-\eta_{RET}) \cdot l_i^\alpha}{S_{non\_ret}} & \text{else}
\end{cases} \right.  
\end{align*}

\section{Experiments}
\subsection{Training Dataset}
Primary data sources include bge-en-icl, bge-m3-data, and bge-multilingual-gemma2-data \footnote{\href{https://github.com/FlagOpen/FlagEmbedding/tree/master/dataset}{https://github.com/FlagOpen/FlagEmbedding/tree/master/dataset}} . The E5 dataset (approximately 1.5M samples) \footnote{\href{https://drive.google.com/file/d/1YqgaJIzmBIH37XBxpRPCVzV\_CLh6aOI4/view}{https://drive.google.com/file/d/1YqgaJIzmBIH37XBxpRPCVzV\_CLh6aOI4/view}}, utilized in E5-Mistral-7B\cite{ref28}, Echo Embedding\cite{ref11}, and LLM2Vec\cite{ref111}, is also incorporated. The aforementioned datasets include commonly used retrieval training corpora such as MS MARCO (both passage and document versions)\cite{ref65}, Natural Questions (NQ)\cite{ref66}, ELI5\cite{ref67}, HotpotQA\cite{ref68}, MIRACL\cite{ref69}, SQuAD\cite{ref70}, FEVER\cite{ref71}, Quora Question Pairs(QQP), and DuReader\cite{ref72}, etc. Previous researchers have already systematically collected and organized these datasets, making them readily usable, we solely utilized the proposed method to update harder negative samples. Stella’s\cite{ref53} retrieval\_data\_llm \footnote{\href{https://huggingface.co/datasets/infgrad/retrieval\_data\_llm}{https://huggingface.co/datasets/infgrad/retrieval\_data\_llm}} provides high-quality (query, positive, negative) triplets, while zpoint leverages datasets such as Huatuo medical QA\footnote{\href{https://huggingface.co/iampanda/zpoint\_large\_embedding\_zh}{https://huggingface.co/iampanda/zpoint\_large\_embedding\_zh}}, all above data has been incorporated. Additional data from huggingface’s sentence-transformers\footnote{\href{https://huggingface.co/sentence-transformers}{https://huggingface.co/sentence-transformers}} repository includes reddit, hover\cite{ref73}, mr-tydi\cite{ref74}, law-gpt, and s2orc\cite{ref75}. Other sources encompass web\_questions, BioASQ\cite{ref54}, cmrc\cite{ref55}, CSL\footnote{\href{https://github.com/ydli-ai/CSL?tab=readme-ov-file}{https://github.com/ydli-ai/CSL?tab=readme-ov-file}}, nli\_for\_simcse (used in SimCSE\cite{ref7} and GTE\cite{ref33}), MLDR\footnote{\href{https://huggingface.co/datasets/Shitao/MLDR}{https://huggingface.co/datasets/Shitao/MLDR}}, GLUE Benchmark\cite{ref56}, Yelp Reviews\cite{ref57} and Weibo Sentiment\footnote{\href{https://github.com/SophonPlus/ChineseNlpCorpus?tab=readme-ov-file}{https://github.com/SophonPlus/ChineseNlpCorpus?tab=readme-ov-file}} training sets. \\\\
We further integrate MTEB evaluation-related datasets like Imdb-Classification\cite{ref59}, MassiveIntent-Classification\cite{ref60}, MassiveScenario-Classification\cite{ref60}, STS12\cite{ref61}, LCQMC\cite{ref62}, PAWSX\cite{ref63}, and STSB\cite{ref64}, we utilized the training split from these datasets with contamination exclusion applied to remove samples highly similar to test sets. \\\\
For data requiring format conversion, we apply the methodologies described in Sention \ref{chpt:c2}. Datasets with limited samples (e.g., subsets of bge and e5 series, Imdb-Classification, STS12, LCQMC) are augmented via Paraphrasing and Augmentation (typically applied to datasets with fewer than 60k samples), we ultimately obtained approximately 5M high-quality training samples through API interfaces. We deduplicate all training sets and filter out samples with low query-pos scores using GTE-Qwen2-7B-Instruct \footnote{\href{https://huggingface.co/Alibaba-NLP/gte-Qwen2-7B-instruct}{https://huggingface.co/Alibaba-NLP/gte-Qwen2-7B-instruct}}. For retrieval data lacking hard negatives, we employ synthetic hard negative generation. Due to API cost constraints, only 30\% of hard negatives are synthetically generated; the remainder are produced using stella-large-zh-v3-1792d\cite{ref53}, with top-10 to top-30 ranked results selected as hard negatives. The final training dataset contains 11M quadruples (query, pos, neg, instruction) in total.

\subsection{Trainset Instructions}
For most training data containing instruction formats, we retain their original contents. For the MTEB training set, we adopt instructions corresponding to its evaluation(consistent with Qwen3-Embedding runtime). For external data lacking instructions (e.g., Huatuo, Reddit, Law-GPT, GLUE), we design task-specific and domain-adaptive instructions. Partial instruction templates are provided in Appendix \ref{app:a2}.

\subsection{Training Details}
As previously mentioned, we adopt a two-stage training approach. For the first-stage retrieval training, we train on all retrieval datasets, with a warm-up step of 300 and a learning rate of 3e-5, the total step of training is 32k. In the second stage, we use all training data, set the learning rate to 2e-5, and train for 8k steps, keeping all other configurations the same as in the first stage. We employ a batch size of 256 for all data using the InfoNCE loss (i.e., retrieval and classification), considering data using the cosent loss (i.e., NLI), due to lower memory consumption from the absence of forward computation for negative samples, the batch size is set to 768.  Across all stages, we employ bfloat16 precision, with 4 hard negative samples and a cosine temperature of 0.02, using Adam optimizer with a weight decay of 0.01. The Data Grouping Strategy remains unchanged between the two stages, except that the second stage incorporates all data with a global retrieval ratio $\eta_{RET}$ of 0.72. Unlike existing works that commonly use LoRA fine-tuning, we employ full-parameter fine-tuning at all stages to ensure maximum performance improvement. The query and passage lengths are set to 256 and 1536 respectively. However, in practice, the model can handle sequences up to 8k in length due to the strong length extrapolation capability of the RoPE\cite{ref35} positional encoding used in most LLMs. The hyperparameter configurations for all training stages are provided in the table \ref{tab:t2}.

\begin{table}[H]
\centering
\caption{Training Hyperparameter Specifications}
\label{tab:t2}
{\centering
\begin{tabular}{|c|c|c|}
\hline
\rowcolor{gray!30}
\textbf{Item} & \textbf{Stage1} & \textbf{Stage2} \\
\hline
Warm-up & \multicolumn{2}{|c|}{300} \\
\hline
Steps & 3e-5 & 2e-5 \\
\hline
LR & 32k & 8k \\
\hline
Batch Size InfoNCE & \multicolumn{2}{|c|}{256}  \\
\hline
Batch Size Cosent & - & 768 \\
\hline
Precision & \multicolumn{2}{|c|}{bfloat16} \\
\hline
Temperature & \multicolumn{2}{|c|}{0.02} \\
\hline
Optimizer & \multicolumn{2}{|c|}{Adam} \\
\hline
Query Length & \multicolumn{2}{|c|}{256} \\
\hline
Passage Length & \multicolumn{2}{|c|}{1536} \\
\hline
\end{tabular}\par
}
\end{table}

\subsection{Compared Methods}
We selected the top-10 ranked models(August 27, 2025) on the MTEB/CMTEB leaderboards prior to the release of QZhou-Embedding as baselines. For MTEB, the comparative models include LGAI-Embedding-Preview\cite{ref16}, the Seed series (v1.5\cite{ref76} , v1.6\cite{ref38}), Qwen series (8B, 4B)\cite{ref34}, ritrieve\_zh\_v1, xiaobu-embedding-v2, gemini-embedding-001\cite{ref77}, jasper\_en\_vision\_language\_v1\cite{ref13}, Linq-Embed-Mistral\cite{ref52}, SFR-Embedding-Mistral\cite{ref30}, and NV-Embed-v2\cite{ref47}. For CMTEB, the baseline models comprise the Seed series (as above), Qwen series (as above), Conan series (v1\cite{ref23}, v2\cite{ref12}), zpoint\_large\_embedding\_zh, and piccolo-large-zh-v2\cite{ref39}.

\subsection{Main Results}
This section presents the evaluation results of Qzhou-embedding on MTEB/CMTEB benchmarks, alongside comparative scores from the top 10 ranked models. As detailed in Table \ref{tab:t3}, Table \ref{tab:t4}, Qzhou-embedding achieves state-of-the-art performance across both task-level and task-type average metrics, demonstrating the effectiveness of our approach. Furthermore, under MTEB's official ranking protocol, Qzhou-embedding secured the top position on both leaderboards. (\textbf{Note}: Highlighted maximum values in certain columns may reflect the best performance among the listed models rather than the overall leaderboard maximum, as exemplified by the MTEB/classification benchmark where the top score does not appear in the top 10 models.)

\begin{table}[H]
\centering
\caption{Performance on MTEB(eng, v2)}
\label{tab:t3}
\resizebox{1.1\textwidth}{!}{
\begin{tabular}{|c|c|c|c|c|c|c|c|c|c|c|c|}
\hline
\rowcolor{gray!30}
\textbf{Model} & \textbf{Class.} & \textbf{Clust.} & \textbf{Pair Class.} & \textbf{Rerank.} & \textbf{STS} & \textbf{Retr.} & \textbf{Summ.} & \textbf{Mean(Task)} & \textbf{Mean(TaskType)} \\
\hline
LGAI-Embedding-Preview & 89.97 & 59.25 & 88.67 & 49.13 & 66.18 & 86.69 & \textbf{38.93} & 74.12 & 68.4 \\
\hline
Seed1.5-Embedding & 89.88 & 60.83 & 87.39 & 50.67 & 67.45 & 87.23 & 36.44 & 74.76 & 68.56 \\
\hline
Qwen3-Embedding-8B & \textbf{90.43} & 58.57 & 87.52 & 51.56 & \textbf{69.44} & 88.58 & 34.83 & 75.22 & 68.71 \\
\hline
Qwen3-Embedding-4B & 89.84 & 57.51 & 87.01 & 50.76 & 68.46 & 88.72 & 34.39 & 74.6 & 68.1 \\
\hline
Seed1.6-embedding& 92.42 & 59.22 & 85.07 & 50.28 & 64.9 & 86.87 & 37.1 & 74.07 & 67.98 \\
\hline
gemini-embedding-001 & 90.05 & 59.39 & 87.7 & 48.59 & 64.35 & 85.29 & 38.28 & 73.3 & 67.67 \\
\hline
jasper\_en\_vision\_language\_v1 & 90.27 & 60.52 & 88.14 & 50 & 56.05 & 84.37 & 37.19 & 71.41 & 66.65 \\
\hline
Linq-Embed-Mistral & 83 & 54.07 & 88.44 & 49.44 & 60.14 & 84.69 & 37.26 & 69.8 & 65.29 \\
\hline
SFR-Embedding-Mistral & 80.47 & 54.93 & 88.59 & 50.15 & 59.33 & 84.77 & 36.32 & 69.31 & 64.94 \\
\hline
NV-Embed-v2 & 87.19 & 47.66 & 88.69 & 49.61 & 62.84 & 83.82 & 35.21 & 69.81 & 65 \\
\hline
QZhou-Embedding(Ours) & 88.97 & \textbf{61.65} & \textbf{92.43} & \textbf{51.77} & 67.12 & \textbf{91.65} & 33.05 & \textbf{75.97} & \textbf{69.52} \\
\hline
\end{tabular}
}
\end{table}

\begin{table}[H]
\centering
\caption{Performance on CMTEB(cmn, v1)}
\label{tab:t4}
\resizebox{1.1\textwidth}{!}{
\begin{tabular}{|c|c|c|c|c|c|c|c|c|c|c|}
\hline
\rowcolor{gray!30}
\textbf{Model} & \textbf{Class.} & \textbf{Clust.} & \textbf{Pair Class.} & \textbf{Rerank.} & \textbf{STS} & \textbf{Retr.} & \textbf{Mean(Task)} & \textbf{Mean(TaskType)} \\
\hline
Seed1.6-embedding & 77.98 & 73.11 & 88.71 & 71.65 & \textbf{79.69} & 68.94 & 75.63 & 76.68 \\
\hline
Seed1.5-Embedding & 79.37 & 71.11 & 89.57 & 70.14 & 79.33 & 66.56 & 74.87 & 76.01 \\
\hline
ritrieve\_zh\_v1 & 76.88 & 66.5 & 85.98 & 72.86 & 76.97 & 63.92 & 72.71 & 73.85 \\
\hline
Conan-embedding-v2 & 76.47 & 68.84 & 92.44 & 74.41 & 78.31 & 65.48 & 74.24 & 75.99 \\
\hline
xiaobu-embedding-v2 & 76.53 & 65.17 & 85.94 & 72.58 & 76.49 & 64.18 & 72.36 & 73.48 \\
\hline
Qwen3-Embedding-8B & 76.97 & \textbf{80.08} & 84.23 & 66.99 & 78.21 & 63.53 & 73.84 & 75 \\
\hline
Conan-embedding-v1 & 76.77 & 66.33 & 85.68 & 72.76 & 76.67 & 63.67 & 72.5 & 73.65 \\
\hline
zpoint\_large\_embedding\_zh & 76.4 & 62.23 & 85.75 & 72.33 & 76.36 & 63.86 & 71.81 & 72.82 \\
\hline
piccolo-large-zh-v2 & 76.42 & 62.16 & 85.22 & 70 & 74.36 & 63.46 & 70.86 & 71.94 \\
\hline
Qwen3-Embedding-4B & 75.46 & 77.89 & 83.34 & 66.05 & 77.03 & 61.26 & 72.27 & 73.51 \\
\hline
QZhou-Embedding(Ours) & \textbf{79.99} & 70.91 & \textbf{95.07} & \textbf{74.85} & 78.80 & \textbf{71.89} & \textbf{76.99} & \textbf{78.58} \\
\hline
\end{tabular}
}
\end{table}

\section{Conclusion}
In this technical report, we present QZhou-Embedding, a general-purpose contextual text embedding model with exceptional text representation capabilities. We designed a unified multi-task framework comprising specialized data transformation and training strategies, effectively enhanced the diversity of training data. To further improve the quality of training data and the model’s generalization capabilities, we developed a data synthesis pipeline leveraging LLM API, incorporating techniques such as Paraphrasing, Augmentation, and Hard negative example generation. We employ a two-stage training strategy comprising initial retrieval-focused training followed by full-task fine-tuning, enabling the embedding model to extend its capabilities based on robust retrieval performance. The model achieves state-of-the-art results on the MTEB and CMTEB benchmarks, ranking first on both leaderboards. Our findings establish that data quality and diversity are pivotal for improving embedding model capabilities. In the future, we will focus on developing multimodal and multilingual embedding models, as well as exploring effective applications of embedding models in agent systems, aiming to integrate cutting-edge technologies to optimize this classical module.

\appendix
\section{Appendix}
\subsection{Framework Constraints}
\label{app:a1}
\begin{table}[H]
\centering
\caption{Specifications of framework constraints}
\label{tab:t1}
\begin{tabular}{|>{\columncolor{blue!15}}>{\raggedright}p{5cm}|p{9cm}|}
\hline
\rowcolor{gray!30}
\textbf{Item} & \textbf{Explanation} \\
\hline
Keep core semantics & Preserving the core semantic content, which is the most critical requirement. \\
\hline
Diversity in morphology, syntax, grammar, tense, rhetoric, etc & Variations in lexical composition, syntactic structure, grammatical rules, and tense usage are permitted. \\
\hline
Length within±15\% & The length deviation from the original sentence should not exceed 15\%. \\
\hline
Keep language & The language used must be consistent with the original sentence. \\
\hline
Close in field & The content must remain strictly aligned with the domain of the given sentence. \\
\hline
Topic transfer, expansion, extension, prohibiting pure rewriting & Topic shifting, extension, or elaboration is permitted, but purely paraphrased content (identical to the original topic) is prohibited.\\
\hline
POS is the perfect answer(necessary \& sufficient) & Positive examples must be unambiguous and precisely address the query (necessity condition) while containing exclusively relevant content without extraneous information (sufficiency condition). \\
\hline
Hard NEG: Worse than POS: \\ - Semantic deviation (inadequate) \\ - Including irrelevant information(unnecessary) \\ - Different aspects of the same topic & Hard negative examples must exhibit inferior quality compared to positive instances, with noise introduced through three strategies: 1) semantic deviation (failing to accurately address the query), 2) incorporation of irrelevant information, or 3) maintaining the same topic but diverging in aspects. \\
\hline
Imitation: syntax, sentence structure, structural & Generating hard negative examples by emulating the structural and syntactic patterns of the given positive instance is a critical step to maximize discriminative challenge for the model. \\
\hline
\end{tabular}
\end{table}

\subsection{Instruction Examples}
\label{app:a2}
\begin{table}[H]
\centering
\caption{Instruction for partial training data}
\label{tab:t1}
\begin{tabular}{|>{\columncolor{blue!15}}>{\raggedright}p{5cm}|p{9cm}|}
\hline
\rowcolor{gray!30}
\textbf{Dataset} & \textbf{Instruction} \\
\hline
Huatuo & Given a medical question, retrieve user replies that best answer the question \\
\hline
Reddit & Retrieve the paragraph most semantically similar to the given statement \\
\hline
Law-GPT & Retrieve relevant legal provisions or interpretations for the given case \\
\hline
MNLI/SNLI & Retrieve semantically similar text \\
\hline
Yelp & Classify the customer review of businesses \\
\hline
Weibo & Classify the sentiment of Weibo comments \\
\hline
\end{tabular}
\end{table}

\subsection{Data Synthesis Examples}
\textbf{Note}: The text highlighted in yellow represents the original sentence, followed by the synthetically generated sentence.
\label{app:a3}
\begin{table}[H]
\centering
\caption{Paraphrasing Example (1)}
\begin{tabular}{|>{\raggedright}p{5cm}|p{9cm}|}
\hline
\rowcolor{gray!30}
\textbf{query} & \textbf{pos} \\
\hline
\rowcolor{yellow!15}
What is the best credit card for someone with no credit history? & If you've never had a credit card before a likely reason can be due to lack of credit history. You can apply for a department store card. \\
\hline
What’s the ideal credit card for a person without any credit history? & If you’ve never had a credit card, it’s probably because you don’t have a credit history. A department store card could be a good option to apply for. \\
\hline
What’s the top credit card choice for someone who has no credit history? & If you’ve never owned a credit card, it’s probably because you don’t have a credit history. A department store card might be a good option to consider. \\
\hline
\end{tabular}
\end{table}

\begin{table}[H]
\centering
\caption{Paraphrasing Example (2)}
\begin{tabular}{|>{\raggedright}p{5cm}|p{9cm}|}
\hline
\rowcolor{gray!30}
\textbf{query} & \textbf{pos} \\
\hline
\rowcolor{yellow!15}
Which English Poet Laureate wrote 'The Faerie Queene'? & English Renaissance to begin, shakily, in the 1520s, and it continued until perhaps 1620. England had a strong tradition of literature in the English vernacular, which gradually increased as English use of the printing press became common during the mid 16th century. By the time of Elizabethan literature a vigorous literary culture in both drama and poetry included poets such as Edmund Spenser, whose verse epic 'The Faerie Queene' had a strong influence on English literature but was eventually overshadowed by the lyrics of William Shakespeare, Thomas Wyatt and others. Typically, the works of these playwrights and poets circulated in manuscript form. \\
\hline
Who was the English Poet Laureate that penned 'The Faerie Queene'? & The English Renaissance began, albeit slowly, in the 1520s and likely lasted until around 1620. England boasted a robust literary tradition in the English language, which expanded as the printing press became more widely adopted during the mid-16th century. By the time of Elizabethan literature, a dynamic literary scene in both drama and poetry emerged, featuring writers like Edmund Spenser, whose epic poem The Faerie Queene significantly influenced English literature, although it was eventually overshadowed by the works of William Shakespeare, Thomas Wyatt, and others. Generally, the writings of these poets and playwrights circulated in manuscript form. \\
\hline
Which English Poet Laureate authored 'The Faerie Queene'? & The English Renaissance began somewhat uncertainly in the 1520s, continuing through to around 1620. England had a rich literary heritage in its native tongue, which expanded as the printing press became more widely used in the mid-16th century. By the Elizabethan era, a thriving literary culture in both drama and poetry emerged, with poets like Edmund Spenser, whose epic The Faerie Queene profoundly impacted English literature, though it was eventually eclipsed by the works of William Shakespeare, Thomas Wyatt, and others. Generally, the writings of these playwrights and poets were circulated in manuscript form. \\
\hline
\end{tabular}
\end{table}

\begin{table}[H]
\centering
\caption{Augmentation Example}
\begin{tabular}{|>{\raggedright}p{4cm}|p{5cm}|p{5cm}|}
\hline
\rowcolor{gray!30}
\textbf{query} & \textbf{pos} & \textbf{neg} \\
\hline
\rowcolor{yellow!15}
What is the best credit card for someone with no credit history? & If you've never had a credit card before a likely reason can be due to lack of credit history. You can apply for a department store card. & - \\
\hline
Which credit card is easiest to get approved for with bad credit? & For those with poor credit scores, secured credit cards from Capital One or Discover are often the most accessible options since they require a refundable deposit. & Some premium travel cards like Chase Sapphire have great rewards but usually need excellent credit. Store cards may accept lower scores but have high interest rates. \\
\hline
What credit cards can I get as a college student with no credit? & Students without credit history often qualify for starter cards like Discover it Student or Capital One Journey, which are designed to help build credit. & Premium cards like American Express Platinum require good credit history, though some banks offer student accounts with debit cards. \\
\hline
\rowcolor{yellow!15}
Which English Poet Laureate wrote 'The Faerie Queene'? & ...By the time of Elizabethan literature a vigorous literary culture in both drama and poetry included poets such as Edmund Spenser, whose verse epic 'The Faerie Queene' had a strong influence on English literature but was eventually overshadowed by the lyrics of William ... & - \\
\hline
What major epic poem did Edmund Spenser write during Queen Elizabeth's reign? & Edmund Spenser composed 'The Faerie Queene', an allegorical epic poem that became one of the most significant works of Elizabethan literature though later eclipsed by Shakespeare's popularity. & Christopher Marlowe's 'Hero and Leander' was another notable Elizabethan poem, but unlike Spenser's work it wasn't an epic allegory. \\
\hline
Which poet created 'Paradise Lost' during the English Renaissance? & John Milton authored the epic poem 'Paradise Lost' in the 17th century, a monumental work that explored biblical themes through blank verse and became a cornerstone of English literature. & William Blake's 'The Marriage of Heaven and Hell' also dealt with religious themes, though it was more prophetic than epic in style compared to Milton's masterpiece. \\
\hline
\end{tabular}
\end{table}

\begin{table}[H]
\centering
\caption{Hard-Negative Generation Example}
\begin{tabular}{|>{\raggedright}p{4cm}|p{5cm}|p{5cm}|}
\hline
\rowcolor{gray!30}
\textbf{query} & \textbf{pos} & \textbf{neg} \\
\hline
\rowcolor{yellow!15}
What territory was King Hussein afraid Israel would obtain? & ...Hussein was nonetheless wary that an Egyptian-Israeli war would risk the West Bank's occupation by Israel... & - \\
\hline
What territory was King Hussein afraid Israel would obtain? & ...Hussein was nonetheless wary that an Egyptian-Israeli war would risk the West Bank's occupation by Israel... & King Hussein expressed concerns about potential Israeli expansion during the Arab-Israeli conflicts, though his warnings to Nasser were delayed and initially dismissed, while other Arab leaders focused more on direct military preparations against Israel. \\
\hline
What territory was King Hussein afraid Israel would obtain? & ...Hussein was nonetheless wary that an Egyptian-Israeli war would risk the West Bank's occupation by Israel... & King Hussein expressed concerns about potential Israeli territorial expansion during the 1967 tensions, though his warnings were delayed in reaching Nasser and mixed with broader regional tensions, while Egyptian military movements in Sinai were already underway under Amer's orders. \\
\hline
\end{tabular}
\end{table}

\end{document}